\documentclass[letterpaper, 10 pt, conference]{ieeeconf}  

\IEEEoverridecommandlockouts                              

\usepackage{graphics} 
\usepackage{epsfig} 
\usepackage{times} 
\usepackage{amsmath} 
\usepackage{amssymb}  

\usepackage{amsfonts}
\usepackage{multirow}
\usepackage{booktabs}
\usepackage{graphicx}
\usepackage{subfigure}
\usepackage{mwe}
\usepackage{threeparttable}

\usepackage{cite}
\usepackage{hyperref}
\hypersetup{
    colorlinks=true,          
    linkcolor=blue,           
    citecolor=blue,           
    urlcolor=blue,            
    pdfborderstyle={/S/U/W 1} 
}

\usepackage{color}

\title{\LARGE \bf
Less is More:  Physical-enhanced Radar-Inertial Odometry 
}

\author{Qiucan Huang, Yuchen Liang, Zhijian Qiao, Shaojie Shen, and Huan Yin
\thanks{This work was supported in part by the HKUST Postgraduate Studentship, in part by the HKUST-DJI Joint Innovation Laboratory, and in part by the Hong Kong Center for Construction Robotics (InnoHK center supported by Hong Kong ITC).}
\thanks{The authors are with the Cheng Kar-Shun Robotics Institute, Hong Kong University of Science and Technology, Hong Kong SAR. E-mail: qhuangag@connect.ust.hk, yliangbk@connect.ust.hk, zqiaoac@connect.ust.hk, eeshaojie@ust.hk, eehyin@ust.hk}
\thanks{Corresponding author: Huan Yin}
}

\begin{document}

\maketitle
\thispagestyle{empty}
\pagestyle{empty}

\begin{abstract}
Radar offers the advantage of providing additional physical properties related to observed objects. In this study, we design a physical-enhanced radar-inertial odometry system that capitalizes on the Doppler velocities and radar cross-section information. The filter for static radar points, correspondence estimation, and residual functions are all strengthened by integrating the physical properties. We conduct experiments on both public datasets and our self-collected data, with different mobile platforms and sensor types. Our quantitative results demonstrate that the proposed radar-inertial odometry system outperforms alternative methods using the physical-enhanced components. Our findings also reveal that using the physical properties results in fewer radar points for odometry estimation, but the performance is still guaranteed and even improved, thus aligning with the ``less is more'' principle.
\end{abstract}

\section{Introduction} \label{Introduction}

Radar-based state estimation has a rich history that traces its origins back to the previous century~\cite{clark1999simultaneous}. In recent years, it has become a new trend within the robotics community, primarily driven by significant advancements in addressing challenging conditions~\cite{harlow2023new}. In contrast to the widely adopted visual and laser sensors, radar sensors offer a consistent source of sensing information that remains reliable under all weather conditions, even in adverse scenarios such as dark or foggy indoor environments. This robustness makes radar a dependable sensor, especially for mobile robots operating in demanding and challenging environments.

\begin{figure}[t]
	\centering
	\includegraphics[width=0.9\linewidth] {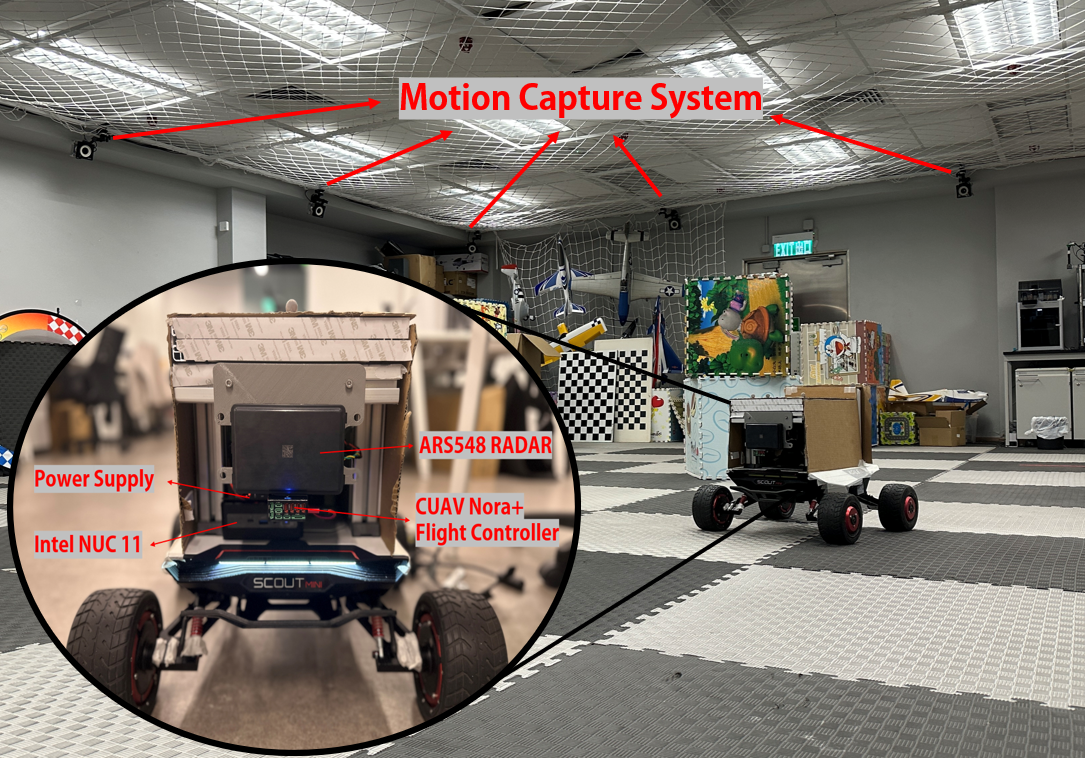}
	\caption{The mobile platform is equipped with a radar sensor and within the experimental room, where a motion capture system is employed to ensure accurate ground truth data for evaluation. }
	\label{figure:platform}
\end{figure}

There is no such thing as a free lunch. The primary challenge arises from radar data being much noisier compared to other sensing modalities, such as data-rich visual images and high-quality laser points. This inherent noise poses significant challenges to radar-based state estimation, especially for invalid or dynamic point elimination and correspondence building with noises. Researchers have proposed several approaches to build a robust radar-based state estimation system. These approaches broadly align with two research directions. The first direction focuses on using novel radar sensors~\cite{barnes2020oxford} or sensing modalities~\cite{cheng2022novel}, designed to deliver dense radar data. It's worth noting that these solutions often come with constraints, such as weight, power supply, and other domain-specific requirements, which restrict their deployment on general robotic platforms, particularly small-sized mobile robots intended for commercial use.

The alternative direction involves the development of estimation systems tailored for conventional radar-on-a-chip sensors~\cite{kramer2020radar,doer2020ekf,zhuang20234d,michalczyk2022tightly,michalczyk2023multi,almalioglu2020milli}. This approach aligns with the advanced methodologies employed in visual/LiDAR-inertial odometry systems (VIO/LIO)~\cite{qin2018vins,xu2022fast}. However, radar points are generally regarded as sparse velocity-aided points in a typical visual or LiDAR matching manner, and the dynamic points and noisy correspondences are not well handled. We consider that a more comprehensive consideration of radar \textit{physical properties} is desired to improve the radar-based state estimation with noises. In this study, we introduce a tightly coupled and physical-enhanced Radar-Inertial Odometry (RIO) method, in which radar's physical properties are leveraged from the front-end radar point filter to the back-end optimization. Specifically, we utilize \textit{Doppler velocities} and inertial measurement unit (IMU) to obtain static radar points, by formulating an IMU-aided velocity check scheme to filter radar points. This approach enhances the system's robustness to noise generated from radar sensors. Additionally, we incorporate \textit{radar cross section} (RCS), also known as radar signatures, to enhance point-to-point correspondence estimation within the proposed RIO system. 

We conduct real-world experiments using both publicly available data and data collected in our motion capture room. Our proposed RIO system stands out by utilizing fewer radar points, a design choice informed by the intrinsic physical properties of radar. Remarkably, as validated in the experiments, using the physical-enhanced filters leads to more accurate metrics compared to those without filters, aligning with the ``less is more'' principle when guided by radar properties. In summary, our specific contributions and findings can be outlined as follows:
\begin{itemize}
    \item We design an IMU-aided velocity check scheme for Doppler velocities and static radar points can be obtained under this scheme.   
    \item We leverage the RCS information to refine point-to-point correspondences, which improves the radar point matching-based pose estimation.
    \item The proposed RIO system is thoroughly evaluated through real-world ablation studies and comparisons, verifying the effectiveness of using radar physical properties in this study.
\end{itemize}






\section{Related Work} \label{Related Works}

The survey paper~\cite{harlow2023new} by Harlow \textit{et al.} reviews the recent radar applications in robotics comprehensively. In this section, we will delve into the research works related to radar(-inertial) odometry.

Several datasets have been made available for Frequency Modulated Continuous Wave (FMCW) Navtech sensors, contributing significantly to the field of 360$^\circ$ radar sensing~\cite{barnes2020oxford,kim2020mulran,sheeny2021radiate}. These datasets have played a pivotal role in advancing radar studies in recent years. Early Navtech radar-based robot odometry primarily relied on traditional visual-inspired keypoints and matching techniques~\cite{cen2018precise,cen2019radar}. Adolfsson \textit{et al.} introduced a high-precision radar odometry approach capable of achieving LiDAR-level precision. Moreover, they developed loop closure detection techniques for global mapping~\cite{adolfsson2022lidar,adolfsson2023tbv}. Advanced machine learning techniques have been applied to enhance radar odometry~\cite{barnes2019masking,burnett2021radar}, rendering the entire framework data-driven and more resilient to noise. In addition to odometry, the Navtech radar-based community has also delved into place recognition~\cite{yin2021radar}, metric localization~\cite{yin2021rall}, and Simultaneous Localization and Mapping (SLAM)~\cite{hong2022radarslam}. However, it's worth noting that Navtech radar sensors have relatively high weight (6 kg) and power requirements (30 W) compared to conventional embedded radar sensors, thereby constraining their application in small-sized robotic platforms.

In recent years, there have been notable contributions from Kramer \textit{et al.}\cite{kramer2022ColoRadar} and Cheng \textit{et al.}\cite{cheng2022novel} in the realm of generating dense sensing using lightweight and low-power radar-on-a-chip sensors, particularly widely-used single or cascaded chips. To achieve dense sensing on such chips, specialized domain knowledge is essential for customizing sensor readings. In the context of mapping, Li \textit{et al.}~\cite{li20234d} proposed a pose graph-based radar mapping using the ZF 4D imaging radar. The 4DRadarSLAM system~\cite{zhang20234dradarslam} used the Oculii Eagle radar sensor, which offers a relatively dense angle resolution, facilitating more precise mapping.

Another research direction involves utilizing radar data produced by cheaper radar chips, with an integration using IMU sensors. For ego-velocity estimation, Kramer \textit{et al.} introduced a method that combines Doppler velocity measurements and inertial measurements through sliding-window optimization~\cite{kramer2020radar}. Doer and Trommer devised an Extended Kalman Filter (EKF)-based radar inertial odometry approach for sensor fusion~\cite{doer2020ekf}. In 4D iRIOM~\cite{zhuang20234d}, Doppler velocity was employed to filter noisy data, and a distribution-to-multi-distribution matching approach inspired by LiDAR was developed for pose estimation~\cite{zhuang20234d}. Michalczyk \textit{et al.} proposed multi-state and tightly-coupled state estimation systems, where radar points served as landmarks for data matching~\cite{michalczyk2022tightly,michalczyk2023multi}. In \cite{almalioglu2020milli}, the radar intensity was considered in the estimation of radar-inertial odometry. To address radar data noisiness, Lu \textit{et al.} introduced a deep learning-based scheme for sensor fusion, thereby enhancing accuracy through a data-driven approach~\cite{lu2020milliego}.

Our radar-inertial odometry system draws inspiration from previous studies within the research community~\cite{kramer2020radar,doer2020ekf,zhuang20234d,michalczyk2022tightly,michalczyk2023multi,almalioglu2020milli}. The difference is the comprehensive utilization of both Doppler velocities and radar signatures (i.e, RCS) to enhance system performance. Our utilization includes front-end data filters, correspondence estimation, and residual functions. The efficacy of the proposed physical-enhanced RIO system has been validated in real-world scenarios, supported by ablation studies to provide further insights into its functionality.

\section{Methodology} \label{Methodology}

\begin{figure*}[t]
	\centering
	\includegraphics[width=16cm]{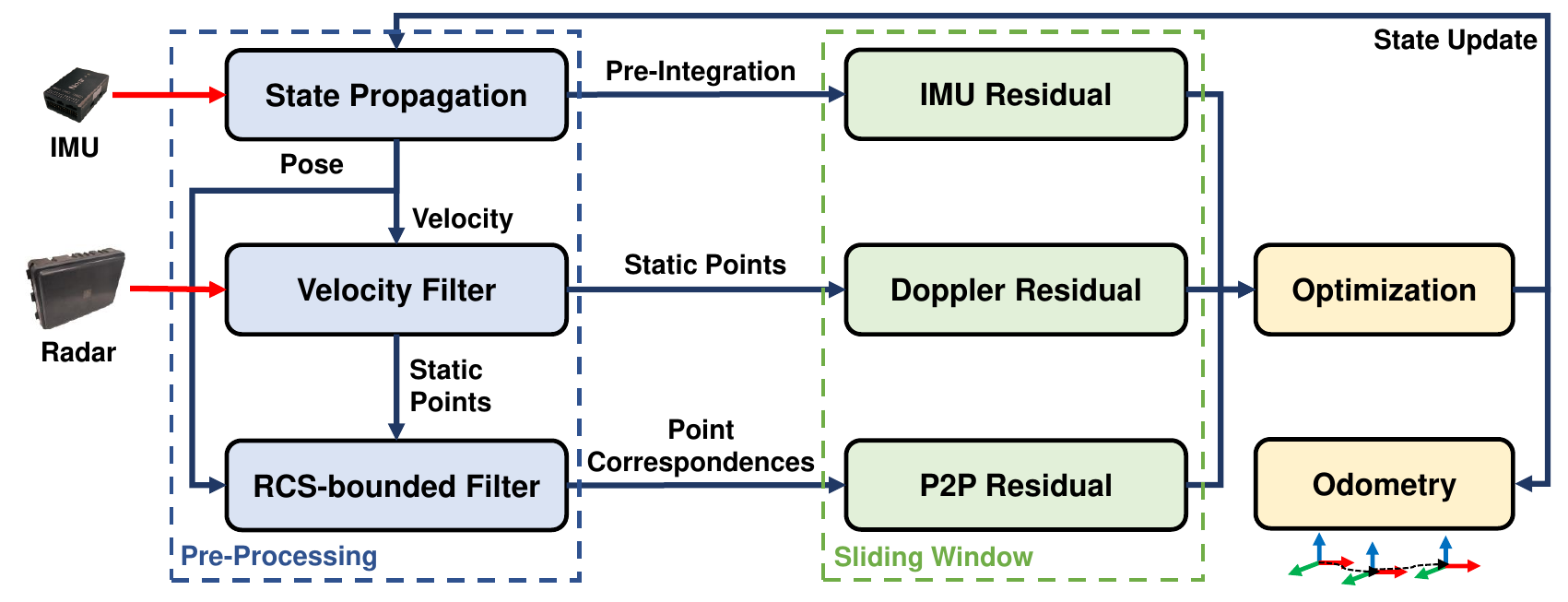}
	\caption{The overview of our proposed physical-enhanced RIO system. The system follows a sliding window-based optimization framework, and with pre-processing modules to handle noisy points and correspondences. Specifically, Doppler velocities are used to obtain static radar points within an IMU-aided velocity check scheme; RCS information is leveraged to bound the neighboring search for point-to-point correspondence estimation. }
	\label{figure:pipeline}
\end{figure*}

\subsection{System Overview} \label{System Overview}

An overview of the system is presented in Figure~\ref{figure:pipeline}, which takes two data as input: sparse radar scans and 6Dof IMU data between two scans. The system is based on a sliding window approach that takes multiple frames to estimate the trajectory of RIO.


The complete state of our system is defined as follows:
\begin{equation}
\begin{aligned}
    &\mathbf{X} = \left[ \mathbf{x}_{0}^{T}, \mathbf{x}_{1}^{T}, \cdots ,\mathbf{x}_{k}^{T}, \mathbf{l}_{0}^{T}, \mathbf{l}_{1}^{T}, \cdots ,\mathbf{l}_{n}^{T}  \right]^{T} \\
    &\mathbf{x}_{i} = \left[ \mathbf{p}_{i}^{T}, \mathbf{q}_{i}^{T}, \mathbf{v}_{i}^{T}, {\mathbf{b}_{a}}_{i}^{T}, {\mathbf{b}_{g}}_{i}^{T}  \right]^{T}, \quad i \in \left[0, k \right] \\
    &\mathbf{l}_{j} \in \mathbb{R}^{3}, \quad j \in \left[ 0, n \right] 
\end{aligned}
\end{equation}
in which $\mathbf{p}_{i}$, $\mathbf{q}_{i}$ and $\mathbf{v}_{i}$ represent the position, rotation, and velocity of the $i$-th frame, respectively.
In the rest of this paper, we use $\mathbf{R}_{i}$ to represent the equivalent rotation matrix of the quaternion $\mathbf{q}_{i}$. The terms ${\mathbf{b}_{a}}_{i}$, ${\mathbf{b}_{g}}_{i}$ denote the accelerometer bias and gyroscope bias in the $i$-th frame, respectively.
Both ${\mathbf{b}_{a}}_{i}$ and ${\mathbf{b}_{g}}_{i}$ are modeled as random walk processes characterized by zero-mean Gaussian noise.
The term $\mathbf{l}_{i}$ refers to the position of the $i$-th landmark in the world frame.

The optimization problem for our system is formulated as follows,
\begin{equation} \label{three-residuals}
\begin{aligned} 
    \min_{\mathbf{X}}\bigg\{ \sum_{k \in I} \lVert\mathbf{r}_{I}(\mathbf{x}_{k},\mathbf{x}_{k+1}) \rVert_{\mathbf{P}_{k}^{k+1}} + & \sum_{k \in D} w_{D} \lVert \mathbf{r}_{D}(\mathbf{x}_{i},p_{k}) \rVert_{2} + \\
    &\sum_{k \in P} w_{P}  \lVert \mathbf{r}_{P}(\mathbf{x}_{i},\mathbf{l}_{k},p_{k}) \rVert_{2} \bigg\}
\end{aligned}
\end{equation}
where $p$ represents a radar point, and $w_{D}$, $w_{P}$ are weights of residuals. The terms $\mathbf{r}_{I}(\cdot)$, $\mathbf{r}_{D}(\cdot)$ and $\mathbf{r}_{P}(\cdot)$ are three different residuals, with further illustrations provided in Section \ref{IMU Preintegration}, Section \ref{Doppler Velocity}, and Section \ref{RCS Corresponding}, respectively.


\subsection{IMU Pre-integration} \label{IMU Preintegration}

Our system utilizes IMU data propagation to estimate continuous radar states. When there are changes in the previous states, re-propagation becomes necessary. However, conducting re-propagation across the entire sliding window can be computationally time-consuming. To circumvent the need for re-propagation, we employ IMU pre-integration, a technique that allows us to streamline the process, as introduced in \cite{qin2018vins}. By employing pre-integration, the residual of IMU measurement is defined as:
\begin{equation}
\begin{aligned}
    &\mathbf{r}_{I}(\mathbf{x}_{k},\mathbf{x}_{k+1}) = \\
    &\begin{bmatrix}     
        \mathbf{R}_{k}^{T}(\mathbf{p}_{k+1} - \mathbf{p}_{k} + \frac{1}{2}g \Delta t^{2} - \mathbf{v}_{k}\Delta t) - \Delta \mathbf{p}_{k+1}^{k} \\
        2 [\mathbf{q}_{k}^{-1} \otimes \mathbf{q}_{k+1} \otimes \Delta \mathbf{q}_{k+1}^{k}]_{xyz} \\
        \mathbf{R}_{k}^{T}(\mathbf{v}_{k+1} + g \Delta t - \mathbf{v}_{k}) - \Delta \mathbf{v}_{k+1}^{k} \\
        {\mathbf{b}_{a}}_{k+1} - {\mathbf{b}_{a}}_{k} \\
        {\mathbf{b}_{g}}_{k+1} - {\mathbf{b}_{g}}_{k} \\
    \end{bmatrix}
\end{aligned}
\end{equation}
where $[\cdot]$ extract the vector part of quaternion and the symbol $\otimes$ denotes the Hamilton product; $\Delta \mathbf{p}_{k+1}^{k}$, $\Delta \mathbf{q}_{k+1}^{k}$ and $\Delta \mathbf{v}_{k+1}^{k}$ are results of IMU pre-integration; $\Delta t$ is the time duration between two scans. Gravity, denoted as $g$, is assumed to align with the $z$-axis.

\subsection{Radar Pre-processing} \label{Radar Preprocessing}

Radar scans frequently include a multitude of invalid points owing to multi-path reflection. Integrating the original data without pre-processing could lead to a failure in state estimation. Consequently, it becomes imperative to eliminate these invalid data points or noises.

The proposed initial step involves discarding data points that lie beyond the current radar settings, such as the field of view (FOV), which serves to reduce computational demands. Subsequently, we employ a radius filter as a criterion~\cite{zhuang20234d} for identifying invalid points, written as follows:
\begin{equation}
    \mathbf{P} = \left\{ p_{i} | \text{Num}(p_{j}) > N \ and \ \lVert p_{i} - p_{j} \rVert \le D \right\}
\end{equation}
where $N$ and $D$ are predefined thresholds; $\mathbf{P}$ represents the set of retained points; $p_{j}$ denotes a neighbor of radar point $p_{i}$. We eliminate points that are excessively far from other points by applying this filter, as scattered points are more likely to be invalid in obtained noisy data.

Radar scans occasionally include false points with error velocities, and the presence of such detection can potentially degrade the estimation performance. A natural approach is to use Doppler velocities for object removal. However, considering the sparse property of radar, velocity estimated solely from radar data may prove inaccurate in specific scenarios, such as when a sizable moving object obstructs the radar's view. To tackle this challenge, we design an IMU-aided velocity check scheme to filter false points, i.e., to obtain real static radar points in environments.


Specifically, we calculate the estimated Doppler velocity $v_{est}$ of radar point ${p_{j}}$ as follows:
\begin{equation} \label{dopplerEq}
    v_{est} = \frac{{p_{j}}^{T}}{\lVert p_{j} \rVert} \cdot \mathbf{R}_{E}^{T} ({\mathbf{R}_{i}}^{T} \mathbf{v}_{i} + (\hat{\omega}_{i} - {\mathbf{b}_{g}}_{t})^{\wedge}\mathbf{t}_{E})
\end{equation} 
where ${\mathbf{R}_{i}}$ and $\mathbf{v}_{i}$ correspond to the rotation and velocity, of the $i$-th radar scan, which are obtained through state propagation. The extrinsic transformation from the radar to the IMU is denoted as $\mathbf{R}_{E}$ and $\mathbf{t}_{E}$, respectively. The term ${\mathbf{b}_{g}}_{t}$ corresponds to the IMU gyroscope bias, and $\hat{\omega}_{i}$ denotes the angular velocity measured by the IMU. The operator $( \cdot )^{\wedge}$ represents the skew-symmetric matrix of the vector $\mathbf{v}$, which is defined as:
\begin{equation}
    \mathbf{v} = \left[ v_{1}, v_{2}, v_{3} \right]^{T}, 
    \mathbf{v}^{\wedge} = \begin{bmatrix}
        0       &-v_{3} &v_{2} \\
        v_{3}   &0      &-v_{1} \\
        -v_{2}  &v_{1}  &0
    \end{bmatrix}
\end{equation}

A radar point is recognized as an inlier if the 2-norm error between the measured Doppler velocity and the estimated Doppler velocity point is within a small threshold, signifying consistency between the two. Then outliers are subsequently excluded from consideration, leaving only the static points with correct velocity measurements for further estimation.

\subsection{Residual on Doppler Velocity} \label{Doppler Velocity}

A radar scan typically includes measurements of azimuth, elevation, range, and Doppler velocity for the detection of targets. In an ideal scenario, the Doppler velocity measured by the radar should align with the projection of the relative velocity between an object and the radar onto the direction from the object to the radar origin.

In the pre-processing step, dynamic points are filtered out with the IMU-aided scheme. With static radar points, the estimation process for the Doppler velocity of reserved static points is described by Equation \ref{dopplerEq}. Then we define the residual of Doppler velocity measurements as follows:
\begin{equation}
\begin{aligned}
    \mathbf{r}_{D}(\mathbf{x}_{i},p_{k}) = \frac{{p_{k}}^{T}}{\lVert p_{k} \rVert} \cdot \mathbf{R}_{E}^{T} ({\mathbf{R}_{i}}^{T} \mathbf{v}_{i} + (\hat{\omega}_{i} - {\mathbf{b}_{g}}_{t})^{\wedge}\mathbf{t}_{E}) - {v_{d}}_{k}
\end{aligned}
\end{equation} 
where ${v_{d}}_{k}$ is the Doppler velocity of $p_{k}$; and $p_{k}$ is the radar measurement in the $i$-th radar scan under a Cartesian coordinate system with the radar origin as its origin point.

\subsection{RCS-bounded Filter for Point-to-Point Residual} \label{RCS Corresponding}

Correspondence plays a pivotal role in point-based matching methods. Previous studies have demonstrated that tracking radar points between subsequent scans is possible. However, there will still be noises or invalid points in the reserved points after filters, the correspondence estimated by searching for the nearest neighbors in Euclidean space might not always be precise. To improve the robustness of matching, we integrate RCS in our approach.


We start with the definition of this physical property. According to \cite{schoffmann2021virtual}, the relationship between received and transmitted power can be described as follows: 
\begin{equation} \label{RadarEq}
    \frac{P_{R}}{P_{E}} = \frac{c G_{E} G_{R} \sigma}{(4\pi)^{2} r^{4} }
\end{equation}
in which $\sigma$ is the value of RCS. $P_{R}$ and $P_{E}$ are received power and transmitted power; $G_{R}$ and $G_{E}$ are the received gain and transmitted gain. The term $c$ is a constant coefficient, and $r$ is the distance between the radar origin to the detected object.

Though the RCS of a simple object can be analytically determined, real-world environments are typically less than ideal, and the robot motions need to be considered. In the field of computer vision, optical flow~\cite{lucas1981iterative} relies on the assumption that pixel intensities of an object remain consistent between consecutive frames. This assumption renders optical flow particularly well-suited for visual-inertial navigation systems, as demonstrated in previous work~\cite{qin2018vins}. Taking inspiration from the principles of optical flow, we assume that slight changes in the pose between radar scans have minimal influence on the RCS, i.e., RCS of an object remains consistent between two consecutive radar scans.


We then conduct correspondence estimation between two consecutive scans guided by our assumption. To achieve this, we first transform all points in the current scan to the coordinate system of the previous scan. When considering a target point $p_{j}$ in these two scans, we design an RCS-bounded nearest neighbor point search, described as follows: 
\begin{equation}
\begin{aligned}
    &\mathbf{P}_{rcs} = \{ p_{i} | \lVert p_{i} - p_{j} \rVert \le D \ and \ \lVert \mathbf{\sigma}_{i} - \mathbf{\sigma}_{j} \rVert \le d \} \\
    &\mathbf{P}_{m} = \{ p_{k} | \forall p_{i} \in \mathbf{P}_{rcs}, p_{k} \in \mathbf{P}_{rcs},
    \lVert p_{i} - p_{j} \rVert \ge \lVert p_{k} - p_{j} \rVert \}
\end{aligned}
\end{equation}
in which $D$ and $d$ are the distance threshold and the RCS threshold; $\sigma_{i}$ and $\sigma_{j}$ represent the RCS values of the detected objects. $\mathbf{P}_{m}$ presents the matched point. Considering that noisy correspondences will still remain after RCS and distance filter, we also employ a landmark selection strategy to make the estimation more robust. To achieve that, we select the closest one as the matched point and only consider the radar points that detected more than a certain time, i.e., only repeatable points in multiple frames are regarded as reliable landmarks for optimization.


Finally, with the estimated correspondences, the final 3D point-to-point (P2P) residual is defined as:
\begin{equation}
    \mathbf{r}_{P}(\mathbf{x}_{i},\mathbf{l}_{k},p_{k}) = \mathbf{l}_{k} - (\mathbf{R}_{i}(\mathbf{R}_{E} p_{k} + \mathbf{t}_{E}) + \mathbf{p}_{i})
\end{equation}
where $\mathbf{l}_{k}$ is the position of the landmark. $p_{k}$ is the measurement of $\mathbf{l}_{k}$ in the $i$-th scan. We solve the entire optimization problem (Equation~\ref{three-residuals}) using the Ceres-Solver \cite{agarwalceressolver2022} with derived Jacobians.



\begin{figure*}[t]
    \centering

    \subfigure[Sequence 1 3D]{	
		\includegraphics[width=4.2cm]{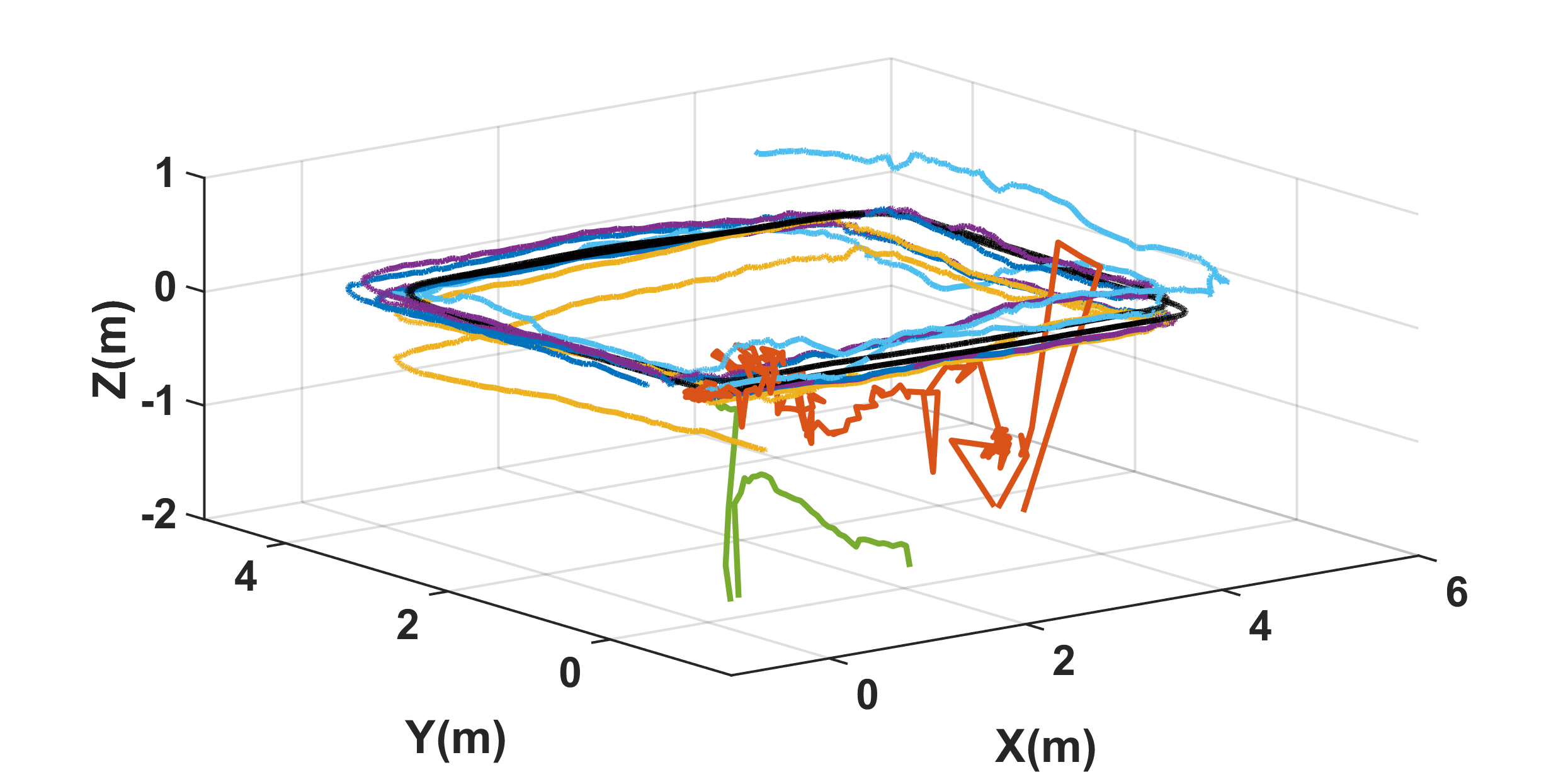}}
    \subfigure[Sequence 1 2D]{	
		\includegraphics[width=4.2cm]{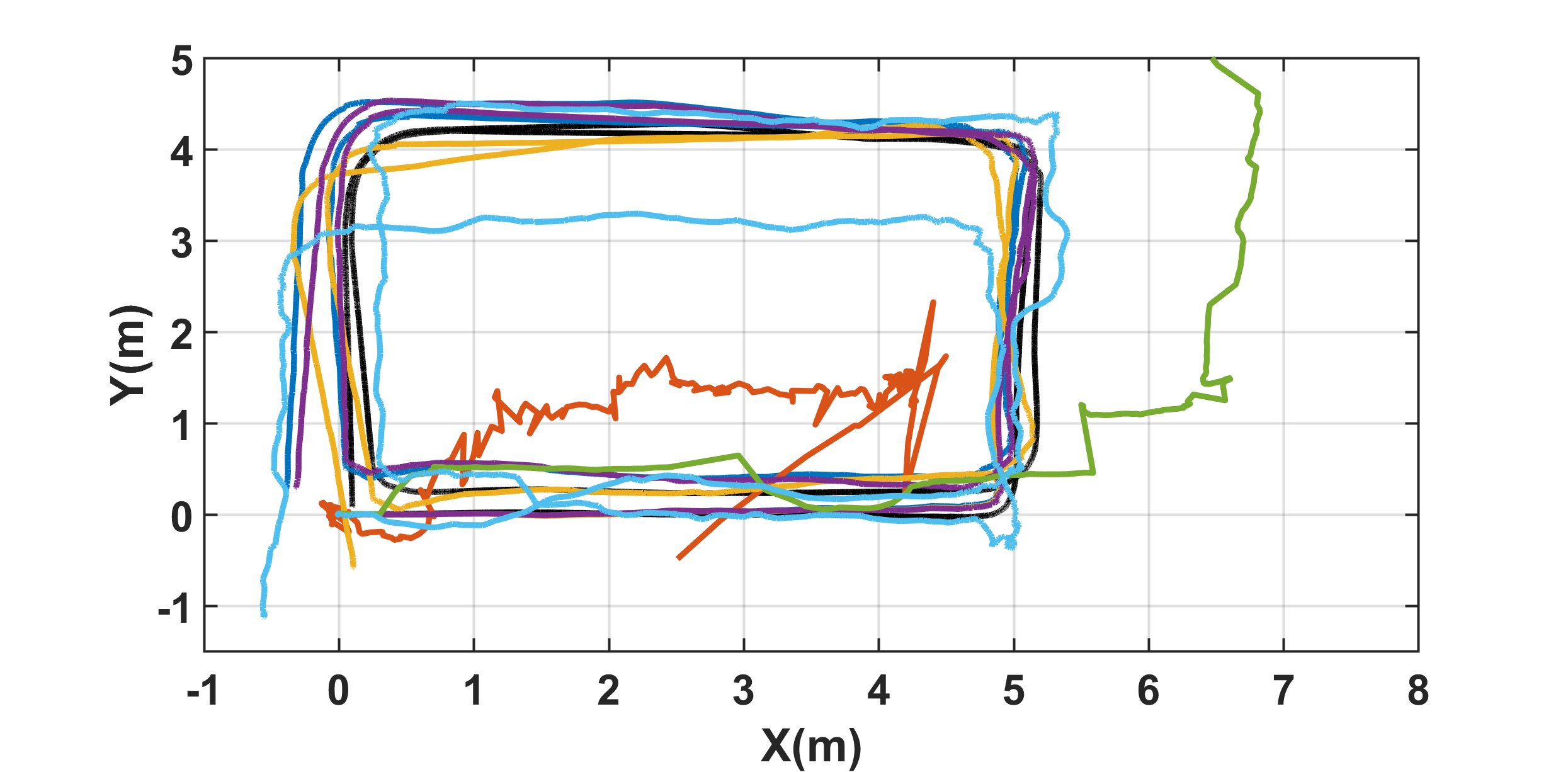}}
  \subfigure[Sequence 2 3D]{	
		\includegraphics[width=4.2cm]{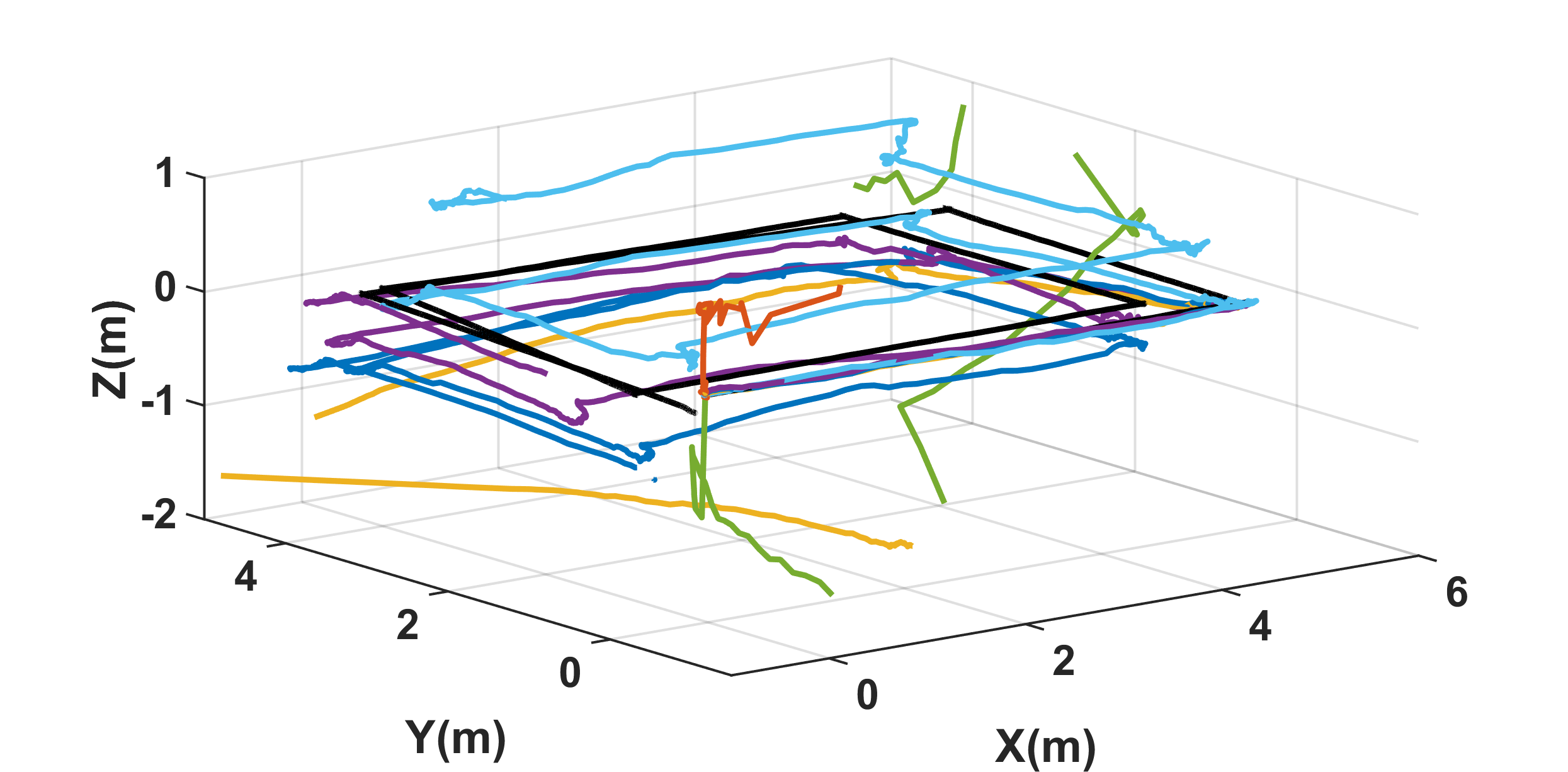}}
  \subfigure[Sequence 2 2D]{	
		\includegraphics[width=4.2cm]{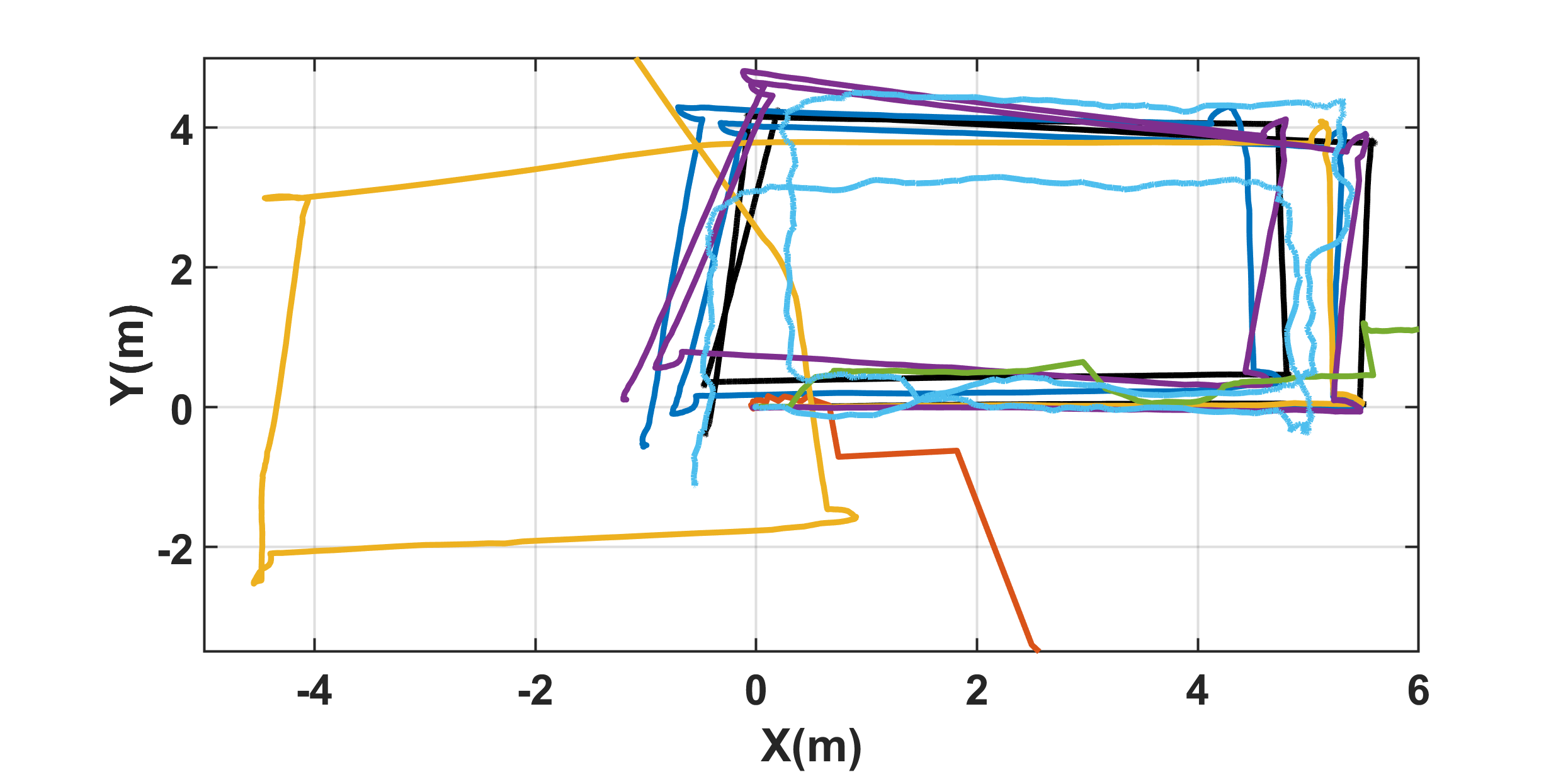}}
  
    \caption[] {Distinct colors are employed to represent different 3D and 2D (x-y) trajectories for ablation study.
    \textcolor[RGB]{0,0,0}{\textbf{Ground Truth}}: obtained from the MoCap system. 
    \textcolor[RGB]{0,115,189}{\textbf{Full System}}: all components enabled. The other ablated variants are listed as follows: 
    \textcolor[RGB]{218,83,25}{\textbf{w/o IMU Residual}},
    \textcolor[RGB]{77,191,239}{\textbf{w/o Velocity Residual}},
    \textcolor[RGB]{238,178,32}{\textbf{w/o P2P Residual}},
    \textcolor[RGB]{119,172,48}{\textbf{w/o Velocity Filter}},
    \textcolor[RGB]{126,47,142}{\textbf{w/o RCS Filter}}, 
    , and all with analyses in Section~\ref{sec:ablation study}.}
    
    \label{trajectoryPlot}
    
\end{figure*}

\section{Experiments} \label{Experiments}

In this section, we first introduce the experimental setup of two datasets. Then quantitative results are presented with ablation studies and comparisons.

\subsection{Setup}

We demonstrate and test our proposed RIO system on a real-world mobile platform, shown in Figure~\ref{figure:platform}, and also a public dataset ColoRadar~\cite{kramer2022ColoRadar}. All the experimental results are evaluated using the open-sourced EVO~\cite{grupp2017evo}.

The mobile platform is a ground robot equipped with ARS548RDI \footnote{https://conti-engineering.com/components/ars-548-rdi/}, a cascaded radar kit from Continental, and an IMU \footnote{https://www.cuav.net/en/nora-plus/}. To generate ground truth poses, we employed a motion capture (MoCap) system. The extrinsic parameters between IMU and Radar are manually calibrated. We collected three distinct sequences for evaluation, each varying in difficulty: Sequence 1 involves relatively low-speed movements; Sequence 2 is collected at higher speeds compared to Sequence 1; Sequence 3 is the most challenging one, characterized by high rotation and translation speeds.




 
We also use the ColoRadar dataset to assess the generalizability of our methods. ColoRadar employs a single-chip Texas Instruments AWR1843BOOST-EVM, and the data is sparser compared to the cascaded radar used in our self-collected dataset. We selected one indoor sequence ``12\_21\_2020\_arpg\_lab\_run0'' (denoted as ColoRadar 1) and one outdoor sequence ``2\_28\_2021\_outdoors\_run0'' (denoted as ColoRadar 2). These two sequences only provide the intensity of each scanned point, which essentially represents the relation between the received and transmitted power. Thus we first remove the distance factor in Equation~\ref{RadarEq} and regard the remaining term as the RCS value. For ColoRadar 1, we employ the MoCap data provided by the dataset as the ground truth. As for ColoRadar 2, the ground truth is obtained from the high-precision LiDAR SLAM system due to the lack of MoCap system. These two sequences were collected using a handheld platform, and aggressive motions bring challenges to the RIO systems.

\begin{figure}[t]
    \centering

    \subfigure[Sequence 1]{	
		\includegraphics[width=4.1cm]{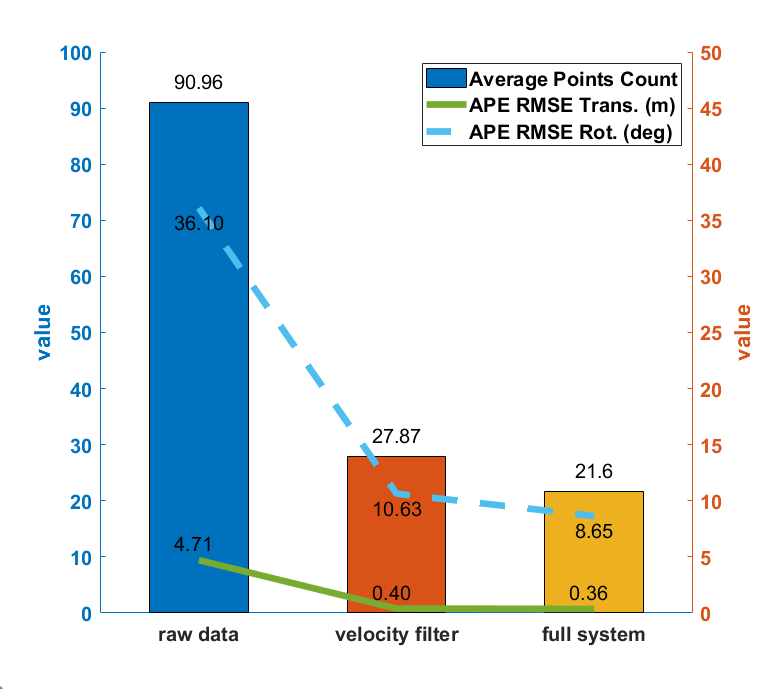}}
    \subfigure[ColoRadar 1]{	
		\includegraphics[width=4.1cm]{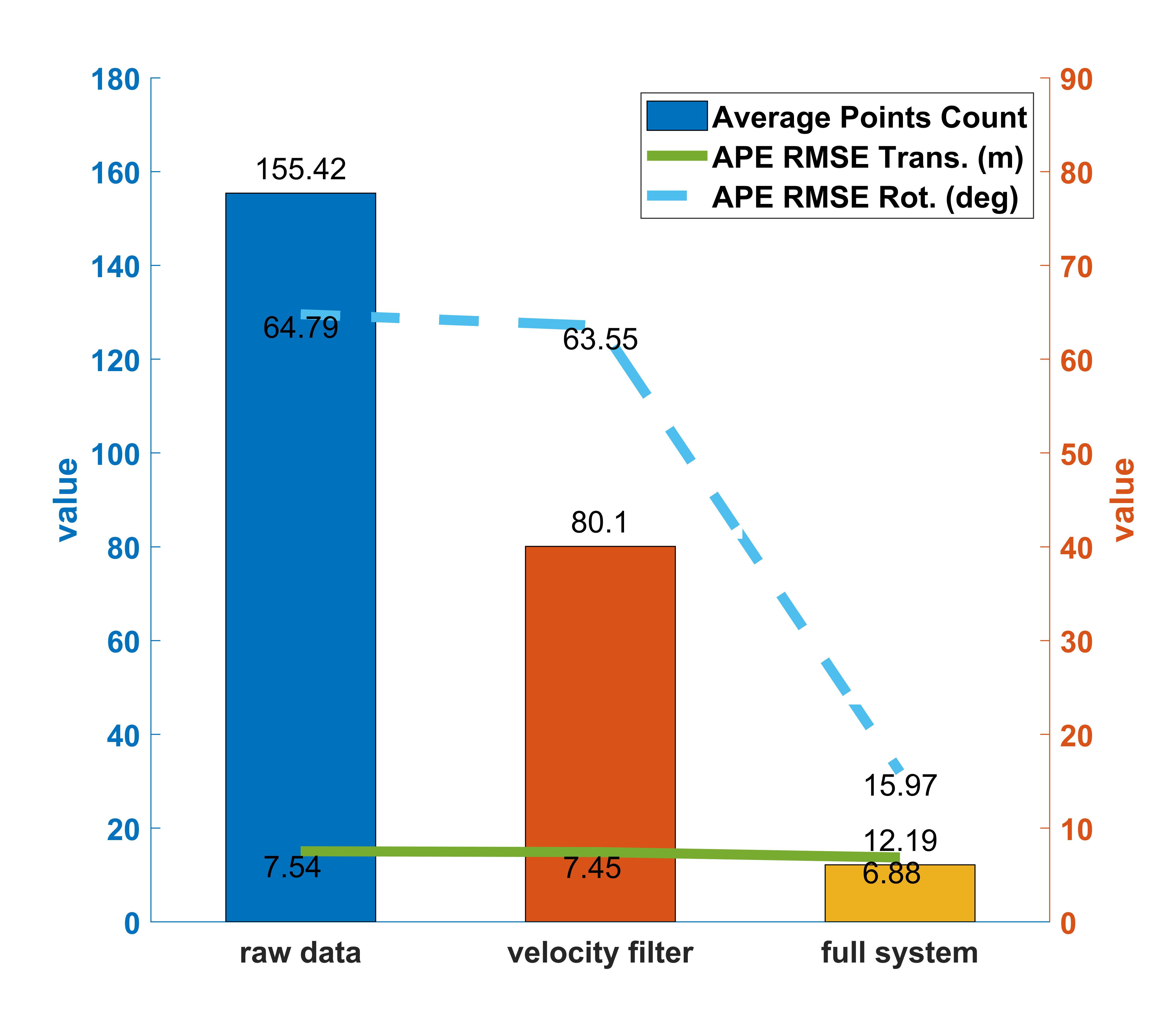}}
    
    \caption[]
    { Average radar point number (left Y-axis) and rotational/positional RMSE (right Y-axis) by applying physical-enhanced filters consecutively: IMU-aided velocity check scheme (denoted as velocity filter) and RCS-bounded filter (denoted as full system).} 
    \label{figure:PC_RMSE}
\end{figure}

\subsection{Less is More}

\begin{figure}[t]
    \centering
        \includegraphics[width=7.5cm]{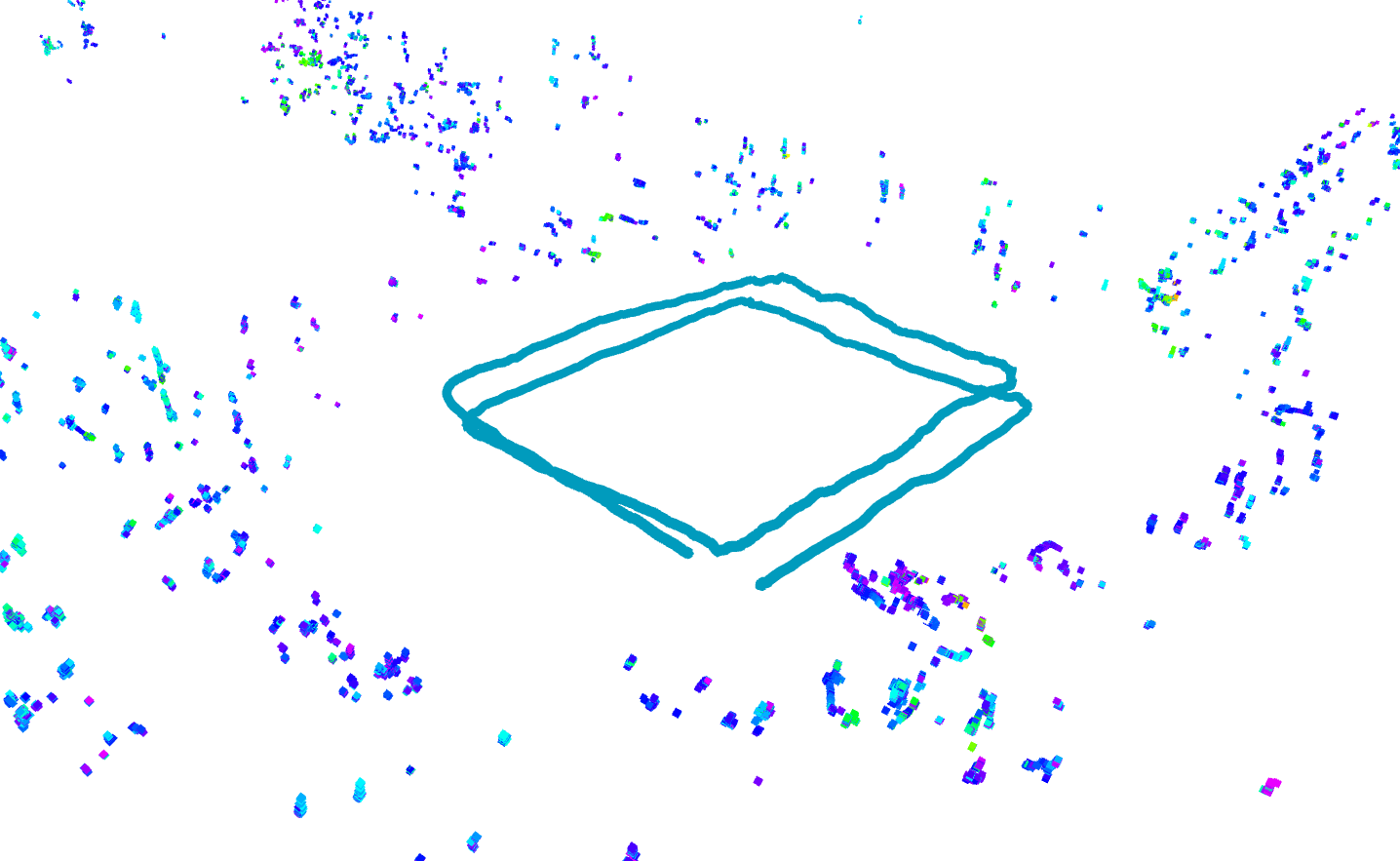}
    \caption[]{Visualization of the radar point cloud map and the traveled trajectory using our proposed RIO. Darker points indicate the radar points with larger RCS values. } 
    \label{mapPlot}
\end{figure}

In the test phase with two datasets, we record the average number of points and the Root Mean Square Error (RMSE) of absolute trajectory error (APE) by employing two filters consecutively: first, the velocity filter to obtain static points, followed by the RCS filter for P2P correspondence estimation. As depicted in Figure~\ref{figure:PC_RMSE}, these two physical-enhanced filters result in a reduction in the radar points utilized for state estimation. Concurrently, RMSE also experiences a decrease when using cleaner radar points, thus underscoring the effectiveness of the ``less is more'' principle in this study. Figure~\ref{mapPlot} presents a visualization of the point cloud map generated from the full RIO system. 

\subsection{Ablation Study} \label{sec:ablation study}

\begin{figure}[t]
    \centering
        \includegraphics[width=\linewidth]{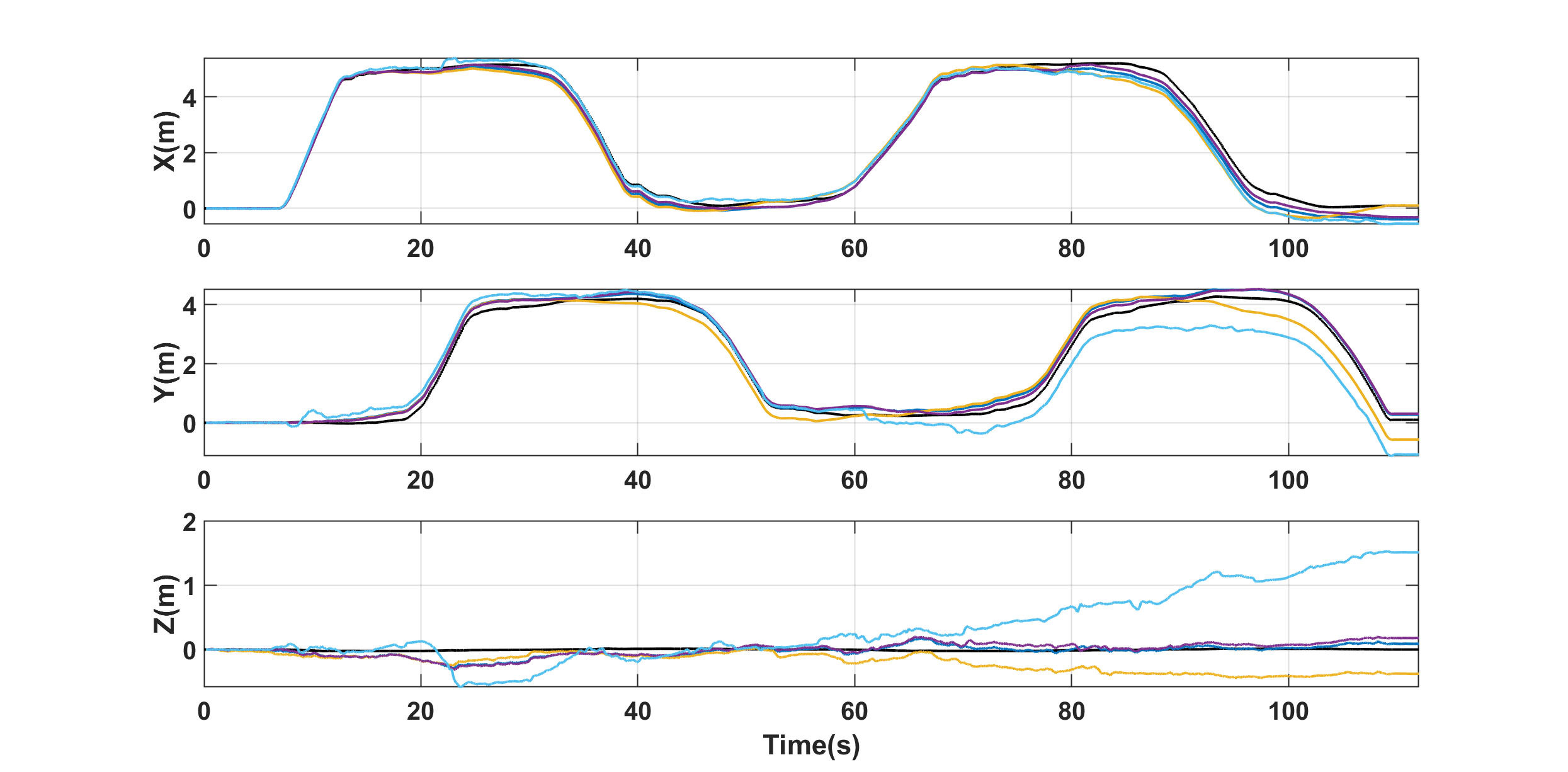}
        \label{sequence1_xyzFig}
    \caption[]
    {Positional errors of Sequence 1 for ablation study. Distinct colors are employed to represent different ablated variants. The color scheme used is consistent with the one depicted in Figure~\ref{trajectoryPlot}.
    
    } 
    \label{PosPlot}
\end{figure}

To assess each component, we performed an ablation study using self-collected Sequence 1 and Sequence 2. We test five ablated variants of our proposed RIO system, each with specific components disabled: IMU Residual, Velocity Residual, P2P Residual, Velocity Residual, Velocity Filter and RCS Filter. The trajectory results of the ablation study are visualized in Figure \ref{trajectoryPlot}, with detailed position information plotted in Figure \ref{PosPlot}. Additionally, the statistics of the RMSE can be found in Table \ref{ablationTB}.

First of all, the full system without any disabled residuals and filters achieves the best performance, meaning that each component contributes to the system. Specifically, we can conclude the following findings from the ablation study:

\textbf{IMU Residual.} The absence of IMU residual $\mathbf{r}_{I}(\cdot)$ results in a rapid accumulation of pose errors. This underscores the fundamental role that the residual $\mathbf{r}_{I}(\cdot)$ plays in the system.

\textbf{Velocity Residual.} If the Doppler velocity residual $\mathbf{r}_{D}(\cdot)$ is disabled, there will be a performance drop for the RIO system. This demonstrates the indispensability of the Doppler property for accurate pose estimation.  

\textbf{P2P Residual.} We find that the point-to-point residual $\mathbf{r}_{P}(\cdot)$ reduces the error when the robot running at a relatively high speed. This indicates that the proposed point-to-point model could improve the estimation with aggressive motions.

\textbf{Velocity Filter.} If the velocity filter is disabled, more dynamic radar points will be involved in the estimation, resulting in a large performance drop for RIO. This indicates the importance of static radar points and validates the effectiveness of our IMU-aided velocity check scheme.

\textbf{RCS Filter.} We also disable the RCS-bounded filter for correspondence estimation, i.e., only distance-based neighbor search is used. As a result, there is a slight performance decrease. We find that its effect becomes prominent when the robot's speed is high, as it could provide more reliable correspondences to stabilize the pose with high speed.



\begin{table}[t] 
\begin{center}   
\caption{Quantitative Results of Ablation Study}  \label{ablationTB} 
\begin{tabular}{cccccc} 
    \hline
        \multirow{2}*{\textbf{Seq.}} &\multirow{2}*{\textbf{method}} & \multicolumn{2}{c}{\textbf{APE RMSE}} & \multicolumn{2}{c}{\textbf{RPE RMSE}} \\
        \cline{3-4} \cline{5-6}
        & &Trans.(m) &Rot.($^{\circ}$) &Trans.(m) &Rot.($^{\circ}$) \\
    \hline
        \multirow{6}*{1} 
        & w/o IMU Res.   & 28.067          & 137.914         & 3.623             & 23.690 \\ 
        & w/o Vel. Res.   & 1.118           & 11.283          & 0.049             & 0.613 \\ 
        & w/o P2P Res. & 0.391           & 8.231           & 0.030             & 0.642 \\ 
        & w/o Vel. Filter   & 6.441           & 64.394          & 0.384             & 11.617 \\
        & w/o RCS Filter   & 0.299           & 8.650           & \textbf{0.027}    & 0.659 \\  
        & Full System    & \textbf{0.294}  & \textbf{8.194}  & \textbf{0.027}    & \textbf{0.599}  \\ 
    \hline
        \multirow{6}*{2} 
        & w/o IMU Res.   & 148.652           & 135.747           & 33.517            & 70.960 \\ 
        & w/o Vel. Res.   & 0.798             & 10.589             & 0.064            & 0.898\\ 
        & w/o P2P Res. & 8.528             & 10.898            & 0.370             & 0.987 \\ 
        & w/o Vel. Filter   & 4.709             & 36.104            & 0.594             & 8.805 \\ 
        & w/o RCS Filter   & 0.396             & 10.635            & 0.062             & 0.901 \\ 
        & Full System    & \textbf{0.358}    & \textbf{8.646}    & \textbf{0.060}    & \textbf{0.879}  \\ 
    \hline
\end{tabular}
\end{center}   
\end{table}

\subsection{Comparisons}

In the following experiments, we compare our method against EKF-RIO~\cite{doer2020ekf}, which is considered one of the best-performing open-source radar-inertial odometry solutions. In addition, we test two alternative approaches for building point correspondences and residual functions:
\begin{itemize}
    \item Point-to-Distribution (P2D): This approach is based on the widely used laser-scan matching techniques. It takes into account the distribution of the current point in relation to the previous map and assigns weights to the distance between the point and the distribution's center, factoring in their covariance.
    \item Distribution-to-Multi-Distribution (D2M) (proposed in \cite{zhuang20234d}): In contrast to the P2D, this approach constructs a distribution for each current point by searching for its neighbors. It then assigns weights to the distance between the center of this distribution and the distributions of its neighbors based on their covariance.
\end{itemize}

\begin{table}[t]    
\begin{center}   
\caption{Quantitative Results of Comparison Study}  \label{compareTB}
\begin{tabular}{cccccc}
\hline
    \multirow{2}*{\textbf{Sequence}} &\multirow{2}*{\textbf{method}} & \multicolumn{2}{c}{\textbf{APE RMSE}} & \multicolumn{2}{c}{\textbf{RPE RMSE}} \\
    \cline{3-4} \cline{5-6}
    & &Trans.(m) &Rot.($^{\circ}$) &Trans.(m) &Rot.($^{\circ}$) \\
\hline
    \multirow{4}*{Sequence 1} 
    & EKF-RIO    & 0.800          & 15.338         & 0.114          & \textbf{0.499} \\ 
    & P2D       & 0.731          & 13.693         & 0.030          & 0.720 \\ 
    & D2M       & 0.565          & 17.007         & 0.030          & 0.640 \\ 
    & Ours      & \textbf{0.294} & \textbf{8.194} & \textbf{0.027} & 0.599 \\ 
\hline
    \multirow{4}*{Sequence 2} 
    & EKF-RIO    & 2.805          & 8.835          & 0.392          & 0.866 \\ 
    & P2D       & 0.701          & 8.943          & \textbf{0.060} & 0.948 \\ 
    & D2M       & 83.841         & 8.883          & 2.065          & \textbf{0.806} \\ 
    & Ours      & \textbf{0.358} & \textbf{8.646} & \textbf{0.060} & 0.879 \\ 
\hline
    \multirow{4}*{Sequence 3} 
    & EKF-RIO    & 68.960         & 66.228         & 2.195          & 2.111 \\ 
    & P2D       & 1.863          & 12.298         & 0.057          & 1.917 \\ 
    & D2M       & 3.180          & 17.717         & 0.274          & 2.408 \\ 
    & Ours      & \textbf{1.098} & \textbf{9.344} & \textbf{0.055} & \textbf{1.619}  \\ 
\hline
    \multirow{4}*{ColoRadar 1}
    & EKF-RIO    &\textbf{5.348}  &24.416          &0.054           &6.044  \\ 
    & P2D       &6.265           &21.837          &0.052           &0.963  \\ 
    & D2M       &6.212           &20.254          &0.057           &\textbf{0.640}  \\ 
    & Ours      &6.877           &\textbf{15.973} &\textbf{0.049}  &0.955  \\ 
\hline
    \multirow{4}*{ColoRadar 2}
    & EKF-RIO    &10.692          &14.764          &0.047            &\textbf{0.497} \\ 
    & P2D       &12.371          &15.913          &0.058               &0.830 \\
    & D2M       &11.076          &\textbf{11.069} &0.054               &0.652 \\
    & Ours      &\textbf{8.789}  &19.578          &\textbf{0.045}      &0.966 \\
\hline
\end{tabular}
\end{center}
\end{table}

We use the open-source version of EKF-RIO and implement the other two methods by ourselves. For the P2D and D2M, we disable our P2P matching and exclude point states during optimization. We also construct submaps to enhance their performance, following the guidelines presented in \cite{zhuang20234d}.

Table \ref{compareTB} presents the RMSE of all comparisons. In summary, our system outperforms other methods when using our self-collected cascaded radar data. However, in the case of the single-chip radar of the ColoRadar dataset, which generates noisier radar measurements, all methods struggle to achieve high accuracy. Specifically, EKF-RIO deals with this issue by discarding all radar measurements, which actually bring drifts in both translation and rotation. As for the P2D and D2M-based methods, these two consider the geometry information around the radar points for matching. However, compared to laser scans, the distribution may not be estimated correctly due to the noise and sparsity of radar data, thus resulting in larger errors of RIO. 

Our proposed RIO filters out unreliable radar points with radar physical properties and tracks the reserved high-quality points in a sliding window. These key points reduce the estimation errors compared to other approaches, therefore our method has better performance in terms of translation since the key points are more stable in translation motion. But on the other hand, the loss of tracking of the same radar points in different radar scans still exists. The rotational motions in the ColoRadar dataset are more aggressive, and the loss of tracking may decrease the performance of methods in terms of rotation. Overall, our method performs at least as well as the aforementioned methods in cascaded millimeter wave radar, but still has some improving space in the future.


\section{Conclusion} \label{Conclusion}

In this study, we design a physical-enhanced radar-inertial odometry system. The physical properties are leveraged to obtain static radar points and build robust pose estimation. The proposed RIO is validated with multiple sequences and two different radar sensor types, and also with comprehensive ablation study analysis and comparisons. Considering the natural properties of radar, there are several promising directions to improve the performance of our RIO system in the future, e.g., integrating graph neural networks for correspondences, and modeling the uncertainty models to weight the residual functions.

\bibliographystyle{IEEEtran}
\bibliography{root}

\end{document}